\documentclass[english]{gretsi}

\usepackage[utf8]{inputenc}
\usepackage[T1]{fontenc}
\usepackage{amsmath, amssymb, amsfonts}
\usepackage{xcolor}

\begin{document}

    \titre{Markov Chain Estimation with In-Context Learning}
    \auteurs{
        \auteur{Simon}{Lepage}
        {s.lepage@criteo.com}{1,2},
        \auteur{Jeremie}{Mary}
        {j.mary@criteo.com}{1}
        \auteur{David}{Picard}
        {david.picard@enpc.fr}{2}
    }
    \affils{
        \affil{1}{CRITEO AI Lab, Paris, France}
        \affil{2}{LIGM, Ecole des Ponts, Univ Gustave Eiffel, CNRS, France}
    }

    \resume{Nous investiguons la capacité des transformers à apprendre des algorithmes utilisant le contexte alors qu'ils sont entraînés uniquement en utilisant la prédiction du prochain token. Nous considérons des chaînes de Markov dont les matrices de transition sont tirées aléatoirement et nous entraînons un transformer a prédire le prochain token. Les matrices d'entraînement et d'évaluation diffèrent et nous montrons qu'à partir d'une certaine taille de modèle et de données d'entraînement, le modèle est capable d'estimer les probabilités de transition à partir du contexte au lieu de mémoriser celles des matrices d'entraînement. De plus, nous montrons qu'encoder astucieusement l'espace d'état permet au modèle de généraliser plus robustement à des chaînes de Markov dont les structures sont différentes de celles vues durant l'apprentissage.}
    \abstract{We investigate the capacity of transformers to learn algorithms involving their context while solely being trained using next token prediction. We set up Markov chains with random transition matrices and we train transformers to predict the next token. Matrices used during training and test are different and we show that there is a threshold in transformer size and in training set size above which the model is able to learn to estimate the transition probabilities from its context instead of memorizing the training patterns. Additionally, we show that more involved encoding of the states enables more robust prediction for Markov chains with structures different than those seen during training.}

    \maketitle

\section{Introduction}

    Transformers, and the attention mechanism at their core, have slowly become one of the most popular neural architectures since their inception as a language modeling tool~\cite{attention_is_all}. Since then, they are now ubiquitous for other tasks such as computer vision~\cite{dosovitskiy2021an}, video generation~\cite{HaCohen2024LTXVideo} or text-to-speech generation~\cite{casanova22yourtts}.
    This success comes largely from the scaling abilities of the transformer architecture which has empirically been shown to handle increases in model size and in training data size very well~\cite{alabdulmohsin22neurips_scaling}.
    In addition, recent works have shown that transformers are universal sequence-to-sequence approximators~\cite{Yun2020Are} and are also Turing complete~\cite{perez21jml_turing} which hints that they should be able to implement complex algorithms by uncovering them from training data.
    
    More recently, large pretrained transformers have been used to perform few-shot tasks by leveraging their context window~\cite{BrownICL2020}. This is called \emph{In Context Learning} (ICL) and assumes that the transformer is able to reason by analogy when pre-pending examples of the same task in the input sequence. 
    In that case, the model is trained with a next-token-prediction loss, that is task agnostic, and is gaining the capabilities to solve new tasks at test time by defining the new task by examples added to the input sequence.
    However, recent work has shown that the ICL mechanism may just be equivalent to a majority vote among the context examples~\cite{Baldassini_2024_CVPRW_ICL}.
    
    In this paper, we want to investigate whether a transformer trained on next-token-prediction can learn to produce an algorithm, and if yes, what are the model sizes and dataset size at which this capacity emerges.
    We set a simple problem of estimating the transition probabilities of a Markov chain.
    Given a (lengthy) realization as input, we want the model to produce a continuation of the sequence that is as likely as possible.
    Of course, we do not want the transformer to memorize the transition probabilities of a single chain, but instead to learn to estimate the transition probabilities of any sequence from the examples given in the context.
    This task would be straightforward if supervised using the empirical estimator of transition probabilities. However, it becomes surprisingly difficult to learn when trained solely with a next-token prediction loss.
    
    Our contributions are the following:
    \begin{itemize}
        \item We thoroughly study the capacity of Transformer architectures to solve next token prediction in Markov Chain using ICL only and show that there is a transition phase between memorizing the training chains and learning to estimate with ICL.
        \item We propose a permutation-based coding scheme that allows us to train on a single Markov chain without losing generalization capabilities.
        \item Finally, we propose a random orthogonal coding scheme that allows us to train with a single structure of Markov chains and generalize to other structures (different number of states, different statistics of valid transitions).
    \end{itemize}

\section{Related Work}
    
    \paragraph{In-Context Learning}
    Synthetic tasks involving sequences sampled from known priors have recently emerged as a powerful approach to studying the emergence of ICL in controlled settings, both theoretically and empirically. In particular, language models have been analyzed through Hidden Markov Models~\cite{xie2022an}, Markov Chains with special "trigger" tokens~\cite{bietti2023birth}, and random finite automata \cite{akyurek2024icl}.
    \cite{edelman2024evolution} examines ICL through Markov Chains and is closely related to our work. However, their study assumes access to unlimited data and does not explore the robustness of the model to changes in the prior.

    \paragraph{Induction Heads}
    The ICL capabilities of language models are largely attributed to attention patterns known as "Induction Heads" - mechanisms that match previous occurrences of the current token and extract information from the subsequent token~\cite{elhage2021mathematical, olsson2022context}.
    In the context of Markov Chains, \cite{edelman2024evolution} demonstrated the emergence of \textit{statistical} induction heads, achieving performance close to that of the Bayes-optimal predictor.

\section{Markov chain estimation}

    We consider the problem of predicting the next token $x_{T+1} \in S$ of a sequence $\mathcal{X} = \{x_1, \dots, x_T\}$, with a state space $S = \{1, \dots, k\}$. We assume that the sequence is a Markov chain with a transition matrix $\mathcal{P}$. More formally, we want to optimize the parameters $\theta$ of a model $f$ such that:
    \begin{align}
        f(\mathcal{X}) \approx \mathcal{P}(x_{T+1} | x_T).
    \end{align}

    Furthermore, we impose a causality constraint on $f(\mathcal{X})$ such that any computation involving $x_t$ can only depend on $x_1, \dots, x_t$. This constraint ensures that the model can subsequently be used in an auto-regressive manner by concatenating a sampling of its output probability distribution to its input. This is a common practice for auto-regressive transformers.
    
    However, instead of trying to learn $\mathcal{P}$ from a large set of realization $\mathcal{X} \sim \mathcal{P}$, we want $f$ to be able to generalize to any Markov chain, including those not seen during training and including those that do not have the same structure (for example, with a different number of states $k$). In some sense, this is a meta learning task as we want the feed-forward call of $f$ to perform online optimization over its input $\mathcal{X}$ so as to predict $x_{T+1}$ (or equivalently $\mathcal{P}$).

    Solving this problem involves 2 key aspects:
    \begin{enumerate}
        \item What training data is used?
        \item What architecture for $f$ is used?
    \end{enumerate}
    We detail both aspects in the following sections.

    \subsection{Data}
        We first consider a simpler case with a fixed number of states $k$. Let us denote by $\mathcal{A} = \{ \mathcal{X} \}$ the training set of size $n$. We consider the case where we have a single training Markov chain, denoted $\mathcal{A}_1 = \{ \mathcal{X}_i \sim \mathcal{P}_1 \}_{1\leq i\leq n}$, and the case where we have $N$ Markov chains, denoted $\mathcal{A}_N = \{ \mathcal{X}_i \sim \mathcal{P}_{i \mod N} \}_{1\leq i\leq n}$, each having their transition matrix sampled according to a Dirichlet distribution ($\mathcal{P}_i \sim \text{Dir}(\boldsymbol{\alpha})$). We chose $\alpha$ small to favor sparse distributions.
    
        For both cases, we sample a test set $\mathcal{T}$ of realizations from 10k different transition matrices to measure the generalization of $f$ to newer Markov chains not seen during training.
    
        Then, we consider a more complex case where the structure of the Markov chain varies between train and test. In particular, we propose to train on a set of realization that have the same Markov chain structure (fixed $k$ and $\alpha$) but evaluate on Markov chains that have a different structure (different $k$ and/or different $\alpha$) to evaluate how $f$ is robust to these changes.

    \subsection{Architecture}
    \paragraph{Transformer with MHSA}

        The core component of the Transformer architecture is the Multi-Head Self-Attention. 
        Given an input embedding $X\in\mathbb{R}^{T\times d}$, and learned projection matrices $W_Q, W_K, W_V\in\mathbb{R}^{d\times d_h}$, where $d_h=d/h$ is the head dimension, each head computes self-attention over $X$. The outputs from all heads are concatenated and projected using $W_O\in\mathbb{R}^{d\times d}$:
        \begin{align}
            Q=XW_Q, \quad K=XW_K, \quad V=XW_V \\
            \text{Attn}(Q, K, V) = \text{softmax}(\frac{QK^\top}{\sqrt{d_h}})V \\
            \text{MHSA}(Q, K, V) = \text{Concat}(\text{head}_1, \ldots, \text{head}_h)W_O
        \end{align}
                
        The decoder blocks follow the Llama architecture~\cite{touvron2023llama}: 
        \begin{align}
            X'_l &= X_l + \text{MHSA}(\text{LN}(X_l)) \\    
            X_{l+1} &= X'_l + \text{MLP}(\text{LN}(X'_l))
        \end{align}
         where MLP is a SwiGLU block~\cite{shazeerglu}, LN is the RMSNorm~\cite{zhang2019root}, and RoPE~\cite{surope} modifies $Q$ and $K$ before attention to encode relative positional information. After the last block, the hidden states are normalized and projected onto the embedding space.

    \subsubsection{State encoding}
    \label{sec:encoding}
    \paragraph{Permutation embedding}
        In the case of $\mathcal{A}_1$, it is then obvious that $f$ will memorize the transitions of $\mathcal{P}_1$ if we are not careful. To prevent this memorization, we propose to relabel the states using a random permutation $\pi$ for each realization. The training set thus becomes $\mathcal{A}_1 = \{ (\pi_i(x_1), \dots, \pi_i(x_T)) | (x_1, \dots, x_T)\sim \mathcal{P}_1 \}$.

    \paragraph{Orthogonal embedding} 
        To avoid being restricted to a fixed space $S$ seen during training, we experiment with dynamically generating the embedding matrix used by the transformer for each sequence. We sample $E \in \mathbb{R}^{k \times d}$ from $\mathcal{N}(0, I)$ and perform a QR decomposition to obtain $E'$ with decorrelated rows. We then compose the input $X$ from $E'$ as $X_i = E'_{x_i}$.

        However, with this approach, we observe that induction heads no longer emerge. We hypothesize that this occurs because the information needed to identify tokens is randomly spread across dimensions, and merging tokens at the first layer of the induction head would destroy critical information.  

        To mitigate this issue, we concatenate a $d$-dimensional vector of zeros to each vector of the vocabulary and increase the network's capacity. At the final layer, we generate logits over the dynamically created vocabulary using a simple dot product.  
    
    \subsubsection{Loss and evaluation}

        During training, we minimize the next-token prediction loss, with $l$ the cross-entropy loss:
        \begin{align}
            L(\theta) = \frac{1}{T}\sum_{i=1}^{T}l(f_\theta(\{x_1\ldots x_i\}), x_{i+1})
        \end{align}
        
        During validation, we compare the predicted transition probabilities against two analytical baselines. The \textit{Oracle}, using the true transition probability matrix $\mathcal{P}$, and the \textit{Empirical estimator}, computed based on the observed context. Denoting $\mathcal{P}^{ij}$ the transition probability from state $i$ to $j$, we derive from our Dirichlet prior that 
        \begin{align}
            \mathbb{E}[\mathcal{P}^{ij}|\mathcal{X}] = \frac{c_{ij}+\alpha}{N_i + k\alpha}
        \end{align} 
        where $c_{ij}$ is the count of transitions from $i$ to $j$ and $N_i$ the number of occurrences of $i$ in the available context. We compare the performance of the estimators using the cross-entropy loss over the last $K=20$ steps. Given $\hat{\mathcal{P}}_{(t)}$ an estimation of $\mathcal{P}$ at step $t$ obtained with the oracle, the empirical estimator, or our model, we report
        \begin{align}
            L(\theta) = \frac{1}{K} \sum_{i=T-K}^T l(\hat{\mathcal{P}}_{(i)}, x_{i+1})
        \end{align}

\section{Experiments}
        
    \subsection{Implementation Details}
        For models with learned embeddings, we use two layers with a single attention head, a hidden size of 128, and tied embeddings. These models are trained for 10k steps. In contrast, models with orthogonal embeddings require additional capacity; we use four layers with two attention heads and train them for 20k steps. Note that applying this configuration to models with learned embeddings does not improve their performances.

        All models are trained using the AdamW optimizer with a constant learning rate of $3\times10^{-4}$, a weight decay of $0.01$, and gradient clipping set to $1.0$. We use a batch size of 500 and train on sequences of length 1000.    

    \subsection{Results}
    \subsubsection{How many $N$ to learn to estimate?}
    
\begin{figure}[t]
            \centering
            \includegraphics[width=\linewidth]{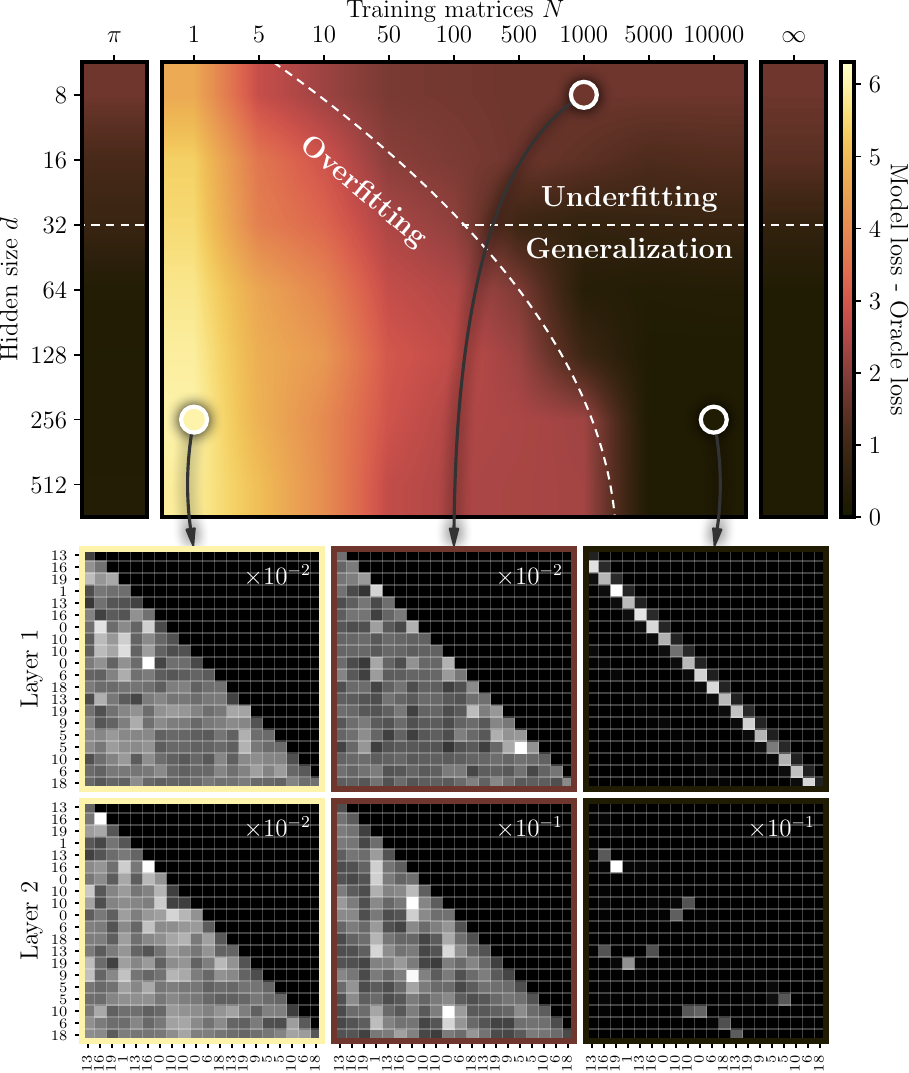}
            \caption{
                Top, performance of models with respect to size and training data. Small models struggle to learn meaningful correlations. Larger models tend to memorize the training priors and do not generalize well for small $N$. Our permutation strategy with a single matrix yields results comparable to training with unlimited sampling.
                For each cell, we report the loss average after 10k training step over 5 runs.
                Bottom, typical attention patterns for each mode.
            }
            \label{fig:D_wrt_K}
        \end{figure}

        We analyze the model's behavior with respect to its size and the number of transition matrices available during training, $N$. We train a collection of models with hidden sizes $d$ ranging from 8 to 512, using 1 to 10,000 predefined Markov chains with $k=30$ and $\alpha=0.1$. We compare their loss to the oracle loss on the test set and present the results in Figure~\ref{fig:D_wrt_K}.

        We observe three distinct regimes:
        \begin{enumerate}
            \item \emph{Overfitting (small $N$)}: The model achieves low training loss while the validation loss increases, especially for large $d$, indicating that it memorizes transition probabilities instead of extracting them from context. The attention is uniform over the entire context.
            \item \emph{Underfitting (large $N$, small $d$)}: Both training and validation losses stagnate, suggesting that the model lacks the capacity to either memorize the training matrices or learn a generalizing algorithm. This behavior persists up to $d=32$, which matches the problem's intrinsic dimensionality for $k=30$. The attention is mostly uniform, with a wavy positional bias at the second layer.
            \item \emph{Generalization (large $d$, large $N$)}: The model suddenly transitions from memorization to generalization, successfully learning to infer unseen transition matrices. The induction head pattern appears.
        \end{enumerate}
        
    \subsubsection{Comparison between $\mathcal{A}_1$ and $\mathcal{A}_N$}

        We focus on the extreme cases: when only a single transition matrix is available for training ($\mathcal{A}_1$), potentially with permutation, and when a new transition matrix $\mathcal{P}$ is sampled for each training sequence (equivalent to $\mathcal{A}_{5\text{M}}$ in our setting).
        These cases are denoted as $\pi$ and $\infty$, respectively, in Figure~\ref{fig:D_wrt_K}.

        While all models quickly overfit on $\mathcal{A}_1$, our simple permutation scheme enables them to generalize, achieving performances comparable to $\mathcal{A}_{5\text{M}}$. This improvement arises from the combinatorial nature of our data augmentation, effectively generating training examples from $k!$ matrices.

        \begin{figure}[t]
            \centering
            \includegraphics[width=\linewidth]{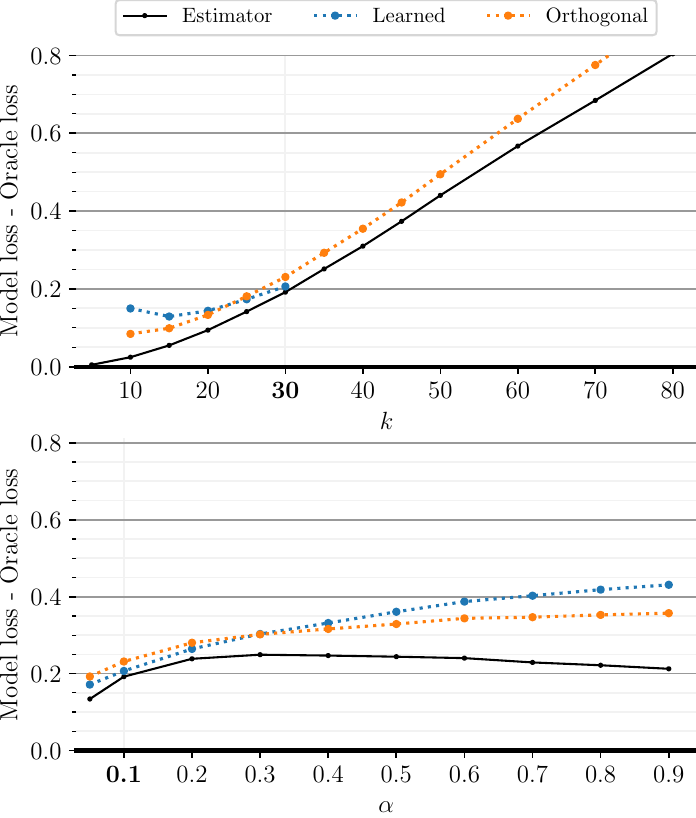}
            \caption{
                Extrapolation capabilities of models trained with $k=30$ and $\alpha=0.1$, for learned and orthogonal embeddings. Top, models performance with respect to the number of states $k$. Models using learned embeddings cannot extrapolate to more states, whereas our model with random orthogonal embeddings follows the trend of the empirical estimator. Bottom, models performance with respect to the prior parameter $\alpha$.
            }
            \label{fig:k-alpha}
        \end{figure}
    
    \subsubsection{Robustness to changes in $k$}

        We now study how differences between the training matrix distribution and the test matrix distribution impact the generalization capability. To that end, we train on random matrices with $k=30$ states and evaluate the performances on matrices with a different number of states, but sampled from Dirichlet distributions of the same parameter $\alpha$. We compare the standard learned embedding of the state with our random orthogonal embedding proposed in \ref{sec:encoding}.
    
        We show the results in Figure~\ref{fig:k-alpha}, top. The standard learned embedding shows degraded performances for Markov chains with fewer states than what was seen during training. Obviously, this model cannot handle Markov chains with a larger number of states since there are no corresponding entries in the embedding matrix. Our proposed orthogonal embedding follows the same behavior as the online estimator, only slightly worse, even for Markov chain more that twice as big as the ones seen during training. This is very encouraging in terms of generalization capabilities as it means that the transformer was able to learn an in-context algorithm.
        
    \subsubsection{Robustness to changes in $\alpha$}

        We now study the robustness to changes of transition statistics by varying the parameter $\alpha$ between training and test. All models are trained with $\alpha=0.1$ and $k=30$.
        Results are shown on Figure~\ref{fig:k-alpha} bottom.
        The online estimator has a relatively constant error, which is not surprising since it incorporates the Dirichlet prior.
        The standard embedding method does really well around the training $\alpha$. But then the error increases significantly has $\alpha$ grows, indicating that the model is perturbed by these entirely new statistics.
        On the contrary, our proposed orthogonal embedding is slightly worse for $\alpha$ close to the training setup, but the the error does not grow as much for larger $\alpha$. 
        This indicates that the orthogonal embedding somehow incentivizes the models to learn a more robust algorithm.

    \section{Conclusion}
    In this paper we show that next-token prediction is a sufficient task to learn to estimate the transition probabilities of a Markov chain from the context instead of memorizing those from the training set, provided that the model size and the training set size are above some threshold.
    We also propose state encoding schemes based on permutations and random orthogonal vectors that allow us to train on a single type of Markov chain while still generalizing to other structures.

     \footnotesize\bibliography{biblio}
\end{document}